\documentclass[sigconf,natbib=false]{acmart}

\usepackage{microtype}
\usepackage{graphicx}
\usepackage{subfigure}
\usepackage{booktabs} 
\usepackage{amsmath} 
\usepackage{xcolor}
\usepackage{tabularx}
\usepackage{algorithmicx}
\usepackage{algorithm}
\usepackage{algpseudocode} 

\newcommand{\R}{\mathbb{R}}

\newcommand{\B}[1]{\textbf{#1}}

\AtBeginDocument{%
  }

\setcopyright{acmlicensed}
\copyrightyear{2024}
\acmYear{2024}
\acmDOI{XXXXXXX.XXXXXXX}

\acmConference{CONSEQUENCES Workshop at ACM RecSys (CONSEQUENCES@RecSys’24)} {October 14, 2024}{Bari, Italy}
\acmISBN{978-1-4503-XXXX-X/18/06}

\begin{document}

\title[Combining Simulator and IS for Tuning Recommenders]{Combining Open-box Simulation and Importance Sampling for Tuning Large-Scale Recommenders}

\author{Kaushal Paneri}
\email{kapaneri@microsoft.com}
\orcid{0000-0002-8785-0723}
\affiliation{%
  \country{}
}

\author{Michael J. Munje}
\authornote{Work done during an internship at Microsoft.}
\email{michaelmunje@utexas.edu}
\orcid{0000-0002-9168-0923}
\affiliation{%
  \country{}
}

\author{Kailash Singh Maurya}
\email{kamaurya@microsoft.com}
\affiliation{%
  \country{}
}

\author{Adith Swaminathan}
\email{adswamin@microsoft.com}
\affiliation{%
  \country{}
}

\author{Yifan Shi}
\email{yifanshi@microsoft.com}
\affiliation{%
  \country{}
}

\renewcommand{\shortauthors}{Paneri et al.}


\received{30 August 2024}

\settopmatter{printacmref=false}

\maketitle
\pagestyle{empty}

\section{Introduction}

Growing scale of recommender systems require extensive tuning to respond to market dynamics and system changes.
Tunable system parameters are often continuous, influence ranking, and consequently key performance indicators (KPIs) such as revenue per thousand impressions (RPM), clicks, and impression yield (IY).
The exploding dimensions of tunable parameters coupled with scale of the system impose significant computational challenges.

\textit{Open-box simulators} \cite{bayir2018genieopenboxcounterfactual} have been successful in parameter tuning in complex systems~\cite{bayir2018genieopenboxcounterfactual}.
By replaying user sessions along with simulated user behaviors in those sessions, 
the open-box simulators provide faithful estimates for KPIs under counterfactual parameters. 
Let $A$ be the parameter space, and $f$ be a KPI (e.g., as estimated by a simulator). For parameter $a \in A$, The objective is to find $\hat{a}$ such that 
\begin{equation}
    \label{eq:SimOpt}
    f(\hat{a}) \approx max_{a \in A} f(a)
\end{equation}
This baseline approach using simulator and finds $\hat{a}$ by enumerating all $a \in A$ (discretizing $A$ if the parameters are continuous). 
For a complex system like for ads recommendation, $f$ can be very expensive to evaluate.
Given $N$ as number of sessions to be replayed (in order of Millions in case of large-scale systems), $s$ be the cost of replaying each session, and $A$ be the number of candidate parameters to be evaluated, the cost is $O(ANs)$.
Stochastic sampling can be used to reduce effective $N$, but when there are plenty of continuous parameters to tune, $A$ can be prohibitively large. 

\textit{Importance sampling} (IS) is another popular approach which offers a cheap surrogate to evaluate many counterfactual parameters efficiently \cite{JMLR:v14:bottou13a,gorurautomated}.
The idea is to randomize the parameters chosen for a subset of traffic, so as to capture the effects of changing the parameters.\footnote{Note that this is different from importance sampling to estimate a black-box function $f(a)$. Specifically, since $f(a)=\frac{1}{N}\sum_i f(a \mid x_i)$ averages over user sessions $x_i$, we can importance sample using sessions, rather than treat $f(a): a \sim A$ as a sample.}
Let $a_0$ be an initial parameter, and $q$ be a distribution imposed to randomize around $a_0$ (e.g. Gaussian), i.e. $a \sim q(a_0)$. Importance sampling returns an unbiased surrogate $\hat{f}(a')=\frac{1}{N} \sum_i f(a_i \mid x_i) \frac{q(a_i \mid a')}{q(a_i \mid a_0)}$ for $a'$ ``close to'' $a_0$~\cite{JMLR:v14:bottou13a,mcbook}.
Fig~\ref{fig:IpsWithSimulation} shows empirically that in our application the IS estimates $\hat{f}$ correlate well with true $f$ evaluations.
IS can therefore be used to maximize 
\begin{equation}
    \label{eq:IterOpt}
\hat{a} = max_{a' \in \{a_0 + \delta\}} \hat{f}(a'),
\end{equation}

where $\{a_0 + \delta \}$ denotes the effective coverage of parameter values through importance sampling (the effective neighborhood $\delta$ depends on $N$, randomization $q$, variability of $f$ across $x_i$, etc.). 
This suggests another practical baseline approach for our original problem:
Initialize $a_0$ and iteratively improve it using $a_{i+1} = arg max_{a' \in \{a_i + \delta\}} \hat{f}(a')$.
We can evaluate the iterates found by IS using the true $f$ to detect progress. For $T$ iterations, the cost of this approach is $O(T*(s + NA_\delta))$. Since $A_\delta$ (the effective number of parameters searched in each iteration) can be much smaller than $A$, and re-weighting samples can be much faster than $O(s)$, this approach is typically computationally efficient compared to enumerating all $a \in A$, however it is sensitive to the choice of $a_0$.

We propose \textit{Simulator-Guided Importance Sampling} (\textbf{SGIS}), which leverages strengths of both simulations and IS.
Other approaches (e.g. doubly robust estimators~\cite{dudik2011doubly}) that combine simulations and IS are motivated by a bias-variance trade-off (offsetting the potential bias of simulator using an unbiased IS alternative); whereas we are motivated to combine two unbiased estimators of KPIs to achieve a computation trade-off.
Using real-world A/B tests, we show that \B{SGIS} driven parameter tuning perform significantly better than open-box simulator, while keeping lower computation cost.  

\section{Method}

Considering $m$ continuous parameters to tune, a parameter setting is a vector in the subspace of $\mathbb{R}^m$.
As this space can be very large, we first generate a grid on every component of the vector to do a coarse search.
Let $\B{C}$ be a coarse grid after partitioning all the parameters.

We use simulator to get KPIs for each grid point using eq. \ref{eq:SimOpt}.
Details about simulator is mentioned in Appendix \ref{Apdx:Simulator}.
We rank parameter settings using an objective function $\mathcal{L}:\R^{m}\to \R$ with score $r$ (Appendix \ref{Apdx:Opt}).
For each of the top $k$ settings ranked by $\mathcal{L}$, simulator generated artificial user sessions are used to perform importance sampling according to eq.~\ref{eq:IterOpt}  (Appendix \ref{Apdx:Art}, \ref{Apdx:IS}).

\begin{algorithm}
    \caption{Simulator-Guided Importance Sampling (SGIS) }\label{algo:SimIPS}
    \begin{algorithmic}[1]
    \Procedure{SGIS}{$\B{X}$,$c$,$d$,$m$,$k$,$u$}
        \State Construct initial coarse grid \B{C}
        \State \B{S} $ = Simulate(\B{X}, \B{C})$ \Comment{Appendix \ref{Apdx:Simulator} Algo \ref{algo:Simulate}}
        \State \B{S} = top-$k$(\B{S}, $\mathcal{L}$, $k$) \label{algo:step:topk1}
        \State Initialize $\B{S}^*$ with $\B{S}$
        \For{$\{1,\hdots,u\}$}
            \For{$[\hat{X}_i, r_i]_{i=1}^{k}$ in $\B{S}$}
                \State $\B{S}_i = ISArt(\hat{X}_i,i)$ \Comment{Appendix \ref{Apdx:IS} Algo \ref{algo:IS}}
            \EndFor
        \State $\B{S} = \bigcup_i \B{S}_{i}$
        \State \B{S} = top-$k$(\B{S}, $\mathcal{L}$, $k$)
        \State Update $\B{S}^*$ with top-$k$ solutions found so far.
        \State Construct a grid \B{C} with settings in $\B{S}$
        \State $\B{S} = Simulate(\B{X}, \B{C})$
        \EndFor
        \State \textbf{return} $\B{S}^*$
    \EndProcedure
    \end{algorithmic}
    \end{algorithm}

\vspace{-10pt}
\begin{figure}[h]
    \centering
    \includegraphics[width=\linewidth]{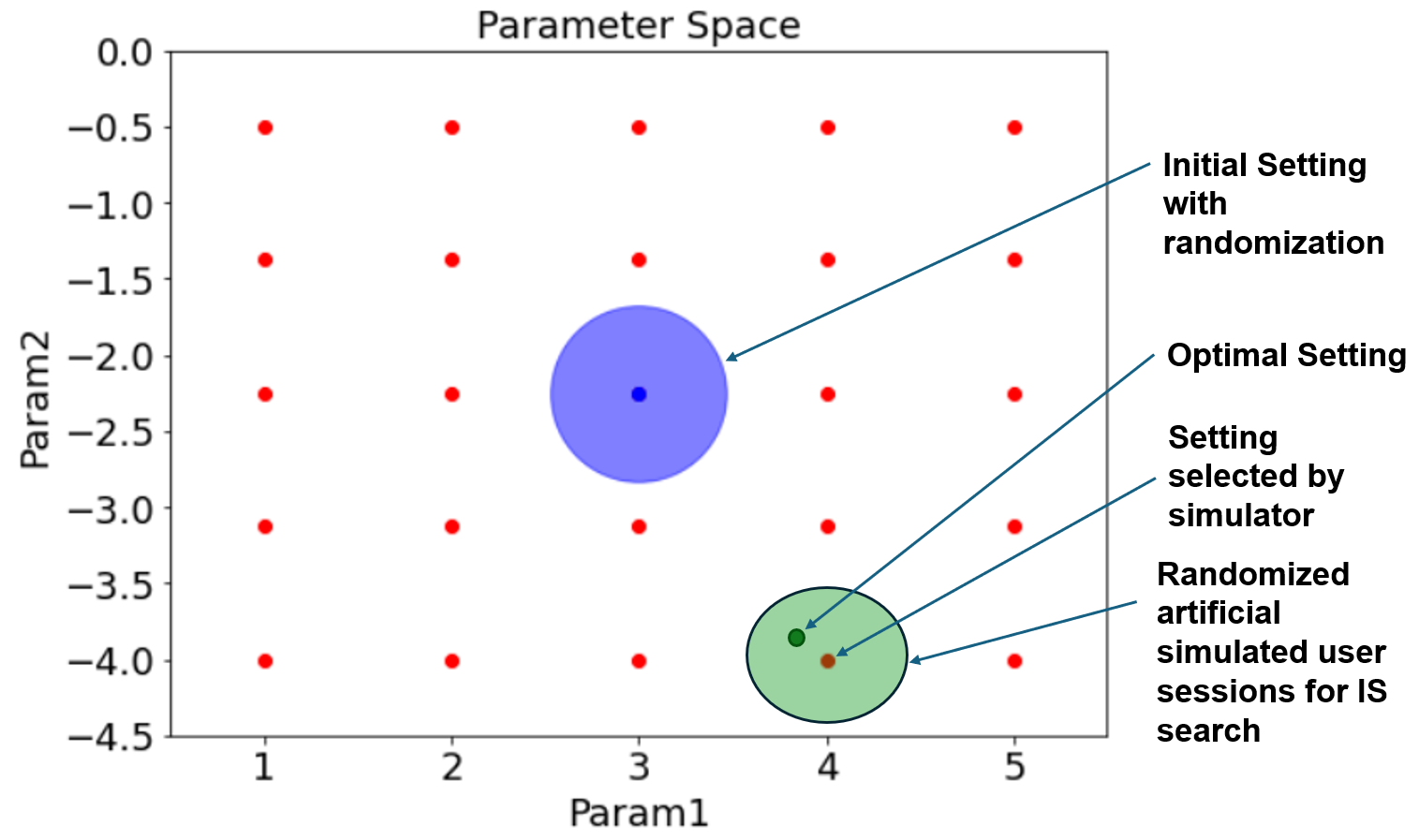}
    \caption{SGIS example in 2-d}
    \label{fig:TwoParamDemo}
\end{figure}

Fig. \ref{fig:TwoParamDemo} demonstrates SGIS with a simple example with two parameters.
The red dots are part of coarse grid evaluated by the simulator, where the blue dot denotes the initial parameter setting.
Let the green dot be the optimal setting in this space, it is possible that the coarse grid will miss this point and will settle with a suboptimal setting.
If we apply IS estimator search with randomization at the initial setting, it will be restricted to the blue region and will need multiple iterations to reach optimal policy (and require collecting randomization data at each iteration).
If the global optimal point is within the range of randomization applied when collecting artificial user session data from simulator, SGIS search with artificial user sessions can reach the optimal point. 
So, our proposed approach only requires that the optimal setting is within the randomization range of at least one of the coarse grid points. 
In Appendix~\ref{Apdx:iter} we discuss the hyper-parameters and trade-offs of our proposal (the coarseness of the grid for enumerative search via  eq.~\ref{eq:SimOpt} vs. the number of iterations for SGIS search. 

\section{Experiments and Results}

For a real-world ad recommendation system, we consider $m=3, c=15, d=25, k=5$. 
Assuming parameters following Gaussian distribution to collect randomized data for SGIS search, the grid for IS search is specified in terms of the sigma multiplier of each parameter.
We select 25 equal distance values from $[-1*\sigma, 1*\sigma]$.
SGIS search is done through a capped-weighted importance sampling estimator (Appendix \ref{Apdx:IS})) \cite{gorurautomated}.
For this experiment, we keep iterations for SGIS search to be $m=1$.
If cost of coarse search is too high, more iterations can be used to reach to an optimal point. (Appendix \ref{Apdx:iter})

\begin{figure}[h]
    \centering
    \includegraphics[width=\linewidth]{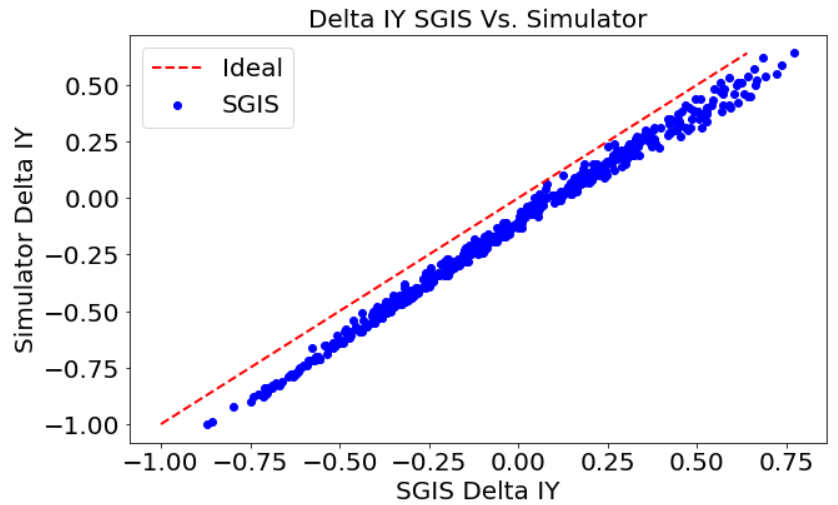}
    \caption{Baseline $\Delta IY$ vs. SGIS $\Delta IY$}
    \label{fig:IpsWithSimulation}
\end{figure}

Fig \ref{fig:IpsWithSimulation} shows SGIS predicted $\Delta IY$ on x-axis and open-box simulator predicted $\Delta IY$ on y-axis on randomly sampled settings from first iteration.
With the high correlation between the two estimators,  
we note that the SGIS estimator can plausibly rank parameters similar to if we had evaluated all settings with the simulator (Eq. \ref{eq:SimOpt}). This observation motivates our Algorithm~\ref{algo:SimIPS}. 
We consider open-box simulations performed for parameter settings with coarse grid in Algo \ref{algo:SimIPS} step \ref{algo:step:topk1} as baseline.

In Table~\ref{tab:ABPerf}, we report the results for parameter optimization when the objective function for ranking candidate settings is $\max_{a} \Delta_a RPM$, s.t. $\Delta_a IY \le 0\%$ (where $\Delta$ denotes differences in KPI relative to a current parameter in deployment).
We pick the best ranking setting obtained from baseline, and compare with the best ranking setting obtained with SGIS.
The $\Delta$ KPI in this table refers to the difference between control and treatment KPIs recorded with A/B experiments ran on a major search engine platform. 

\begin{table}[h!]
\centering
\begin{tabular}{|c|c|c|c|}
\hline
Experiment & Revenue & IY \\
\hline
Simulator Proposed Setting & 0.55\% & -1.67\% \\
\hline
SGIS Proposed Setting & 1.11\% & -0.74\% \\
\hline
\end{tabular}
\caption{A/B Test Performance Comparison}
\label{tab:ABPerf}
\end{table}

\section{Conclusion}

We propose an off-policy evaluator \B{SGIS} that uses expensive simulator to guide a cheap IS search; 
and show using simulations and real world A/B tests that our evaluator lowers the computation cost for off-policy optimization while maintaining performance.
\newpage
\bibliographystyle{acm}  
\bibliography{main}

\newpage
\appendix

\section{Off-policy Estimators}

\subsection{Open-box Simulator}
\label{Apdx:Simulator}

Open-box monte-carlo simulators replaying user sessions are widely used for parameter tuning in complex systems. \cite{bayir2018genieopenboxcounterfactual}.
They use historical user sessions and replay them by running the exact binary that online system runs.
Machine learning models are then used to estimate user response signals.

\begin{figure}[h]
    \centering
    \includegraphics[width=\linewidth]{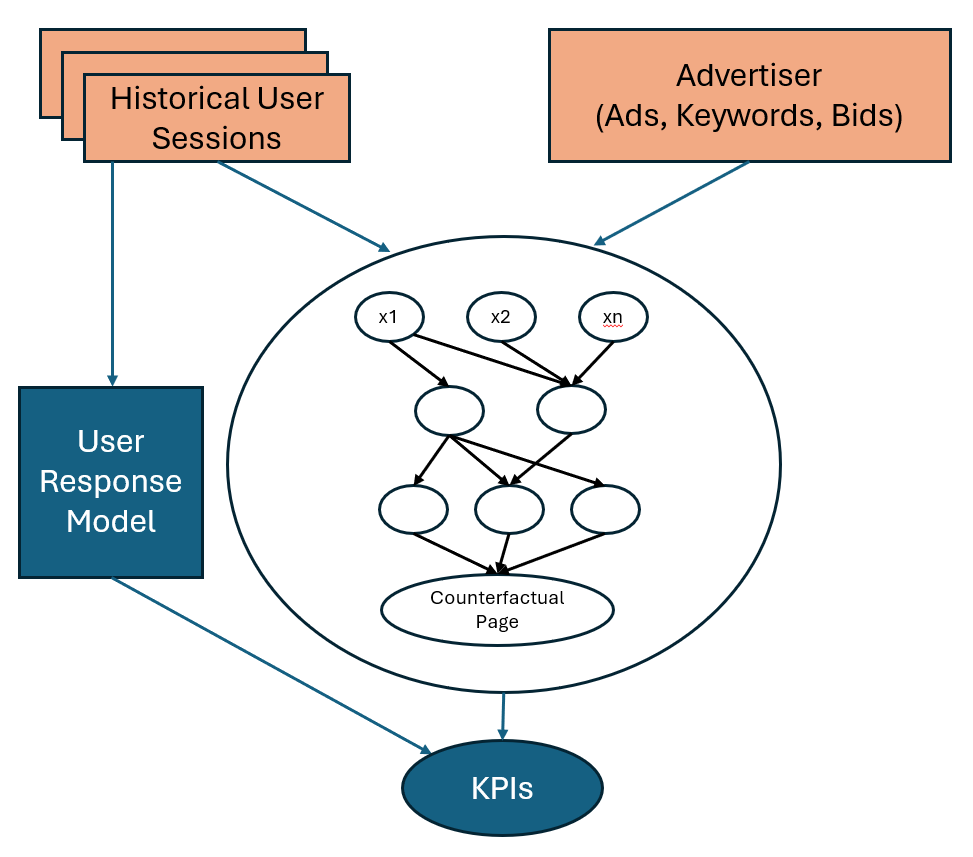}
    \caption{For each parameter setting, simulator can re-run the causal graph in offline fashion to evaluate its effect on KPIs (e.g. RPM, Clicks, IY)}
    \label{fig:Simulator}
\end{figure}

As shown in Fig \ref{fig:Simulator}, given a recommender system laid out as a causal graph that can be replayed by a simulator, combined with a user response model, can be used to evaluate different parameter settings and capture their effect on various KPIs.
As open-box simulations depict the true underlying data generating processes, they provide faithful counterfactual estimates that hold in A/B experiments.

Let $\B{X} = [x_1, \hdots, x_M]$ be historical user sessions.
We first learn a model $\mathcal{M}$ from $\B{X}$ to predict user responses for the simulations.
For each policy $a \in \B{A}$, each user session $x$ is replayed to generate a counterfactual session $\hat{x}$, which in conjunction with the model $\mathcal{M}$ is used to calculate KPIs.
Let $\B{p}$ be a vector containing each KPI, it is aggregated all sessions to produce score vector for each policy.
This score vector can be used to perform multi-objective constrained optimization.

\subsubsection{Artificial User Sessions}
\label{Apdx:Art}

Step \ref{algo:step:uhat} in the algorithm generating a counterfactual user session $\hat{x}$ can be collected for all historical logs in $X$ for a parameter setting $a$.
Considering $a$ has been randomized to capture small region $\delta$, these simulated sessions can be used to perform importance sampling to cheaply evaluate lots of parameter settings.

\begin{algorithm}
    \caption{Simulate}\label{algo:Simulate}
    \begin{algorithmic}[1]
    \Procedure{Simulate}{$\B{X}$, $\B{A}$}
        \State Learn a model $\mathcal{M}$ to predict user responses using $\B{U}$.
        \For{$a$ in $\B{A}$}
            \For{$x$ in $\B{X}$}
            \State Replay user session $x$ with setting $a$ to get $\hat{x}$. \label{algo:step:uhat}
            \State $\B{p}_{xa} = f(\mathcal{M}(\hat{x}), \hat{x})$
            \EndFor
            \State $\B{p}_{a} = \sum_{i=1}^x \B{p}_{xa}$ 
        \EndFor
        \State $\hat{X}_a = [\hat{x}_a]_{\forall a \in \B{A}}$
        \State \B{return} $[\hat{X}_a, p_a]_{\forall a \in \B{A}}$
    \EndProcedure
    \end{algorithmic}
    \end{algorithm}

\subsection{Importance Sampling}
\label{Apdx:IS}

Observational approaches like importance sampling is used for iterative tuning of large-scale recommender systems.
The portion of traffic is randomized to capture the effect of changing parameters.
A popular choice for randomization is imposing a Gaussian distribution \cite{gorurautomated}.
The randomization is controlled with the variance of the proposal distribution $q$, importance sampling is used to evaluate the effect of changing proposal to a candidate distribution $a$.
Let $x$ be a user session and $f$ be a KPI function, the effect of candidate distribution can be captured in the following way.

\begin{equation}
\label{eqn:is}
E_{a}[f(X)] = E_{q}\big[f(X)\frac{a(x)}{q(x)}\big]\approx \frac{1}{N}\sum_{i=1}^{N}f(x_i)\frac{a(x_i)}{q(x_i)}
\end{equation}

As randomization cost can become significant if it is imposed to real traffic, adaptive importance sampling is used in practice to iteratively come up with a better policies to reach an optimal point.

This approach is not restricted to be used only for real observational data.
If we have an accurate simulator representing true underlying processes, we can use it to generate such randomized data, and such iterative importance sampling can be done with no real randomization cost.

We use data collected by simulator for a setting $a$ as explained in \ref{Apdx:Art} to perform importance sampling.

\begin{algorithm}
    \caption{Importance Sampling With Artificial User Sessions}\label{algo:IS}
    \begin{algorithmic}[1]
    \Procedure{ISArt}{\B{$\hat{X}$}, $a$}
        \State Construct a dense grid $A$ around $a\pm \delta$.
        \For{$a$ in $\B{A}$}
            \State $\B{p}_{a} = E_a[f(x)]$ using eq \ref{eqn:is}.
        \EndFor
        \State \B{return} $[\B{p}_a]_{\forall a \in \B{A}}$
    \EndProcedure
    \end{algorithmic}
    \end{algorithm}

\section{Hyperparameter Considerations}
\label{Apdx:Hyper}

\subsection{Top-$k$ Optimization}
\label{Apdx:Opt}
Algorithms for Simulator (\ref{algo:Simulate}) and importance sampling (\ref{algo:IS}) output a vector $\B{p}_a$ containing estimated KPIs like RPM, Clicks, IY for each $a$.
This can be used to construct a multi-objective optimization.
Let $\mathcal{L}:\B{p}\to r$ be an arbitrary objective function to rank all $p$, we can use hyperparameter $k$ to select top-k parameter settings.

KPIs considered for $\mathcal{L}$ influence choice of hyperparameters in algorithm $\ref{algo:SimIPS}$.
In our experiments, IY shows high correlation with simulations as shown in Fig \ref{fig:IpsWithSimulation}.
If $\mathcal{L}$ is constructed only with KPIs that are highly correlated with simualtions, choosing $k=1$ for each iteration works well.
However, KPIs including user-response model predictions like RPM or clicks can have lower correlation.
In such cases, one option is to choose higher $k$ that allows more candidates to be evaluated with simulator at each iteration.
Another option is to collect large number of artificial user sessions to reduce variance of importance sampling.

\subsection{Number Of Iterations for IS Search}
\label{Apdx:iter}

The number of iterations for importance sampling search $u$ is used to tread-off compute cost of simulator and importance sampling estimator. If coarse grid is sparse due to large parameter space (many continuous parameters) or high cost of simulator, we would benefit with big $u$ to reach to an optimal point.
We can also use early stopping if changes in parameter values or KPI gains are less than some $\epsilon$ with each iteration.

Choice of $u$ also depends on randomization applied while collecting simulated user sessions.
As there is no online cost for randomizing simulated user sessions, large randomization can be applied to  reduce the number of iterations $u$.
Further study is required on the impact of large randomization applied to collect simulated user sessions on IS search.

\end{document}